%% file: main.tex
\documentclass[a4paper,fleqn]{cas-sc}

\input{tex/includes}
\input{tex/acronyms}
\input{tex/definitions}

\begin{document}

\input{sections/header} 
\input{sections/abstract}

\maketitle

\input{sections/introduction}
\input{sections/relatedWorks}
\input{sections/referenceMethods}
\input{sections/method}
\input{sections/results}
\input{sections/conclusions}

\section*{Acknowledgements}

This study was partially carried out within the SERICS - SEcurity and RIghts in the CyberSpace and received funding from the European Union Next-Generation EU (Piano Nazionale di Ripresa e Resilienza (PNRR) – Missione 4 Componente 2, Investimento 1.3 – D.D. 1555 11/10/2022, PE7 - CUP J33C22002810001, D.D. 341 15/03/2022, PE\-00000014). This manuscript reflects only the authors’ views and opinions; neither the European Union nor the European Commission can be considered responsible for them.

\printcredits

\bibliographystyle{cas-model2-names} 
\bibliography{biblio/main}

\end{document}

%% file: tex/includes.tex
\usepackage[authoryear]{natbib}

\usepackage[utf8]{inputenc} 
\usepackage[T1]{fontenc} 
\usepackage{color} 
\usepackage{glossaries} 
\usepackage{xspace} 
\usepackage{leftindex} 

\usepackage{graphicx} 
\usepackage{subcaption} 

\usepackage{booktabs} 
\usepackage{tabulary} 
\usepackage{multirow} 

\usepackage{amsmath}  
\usepackage{amssymb}  
\usepackage{amsfonts} 
\usepackage{bm} 

\usepackage[ruled,linesnumbered]{algorithm2e}
\usepackage{lipsum}

%% file: tex/acronyms.tex
\newacronym{aa}{AA}{Adversarial Attack}
\newacronym{ai}{AI}{Artificial Intelligence}
\newacronym{auc}{AUC}{Area Under the Receiver Operating Curve}
\newacronym{apgd}{APGD}{Auto-Projected Gradient Descent}

\newacronym{bim}{BIM}{Basic Iterative Method}

\newacronym{cnn}{CNN}{Convolutional Neural Network}
\newacronym{cl}{CL}{Convolutional Layer}
\newacronym{cam}{CAM}{Coverage Analysis Method}

\newacronym{dnn}{DNN}{Deep Neural Network}
\newacronym{dmd}{DMD}{Deep Mahalanobis Detector}
\newacronym{dr}{DR}{Dimensionality Reduction}

\newacronym{eu}{EU}{European Union}

\newacronym{fab}{FAB}{Fast Adaptive Boundary}

\newacronym{gmm}{GMM}{Gaussian Mixture Model}

\newacronym{hlf}{HLF}{High-Level Feature}

\newacronym{id}{ID}{In-Distribution}
\newacronym{ia}{IA}{Intermediate Activation}
\newacronym{ica}{ICA}{Independent Component Analysis} 

\newacronym{kde}{KDE}{Kernel Density Estimation}
\newacronym{knn}{KNN}{K-Nearest Neighbors}

\newacronym{llf}{LLF}{Low-Level Feature}
\newacronym{llm}{LLM}{Large Language Model}

\newacronym{macs}{MACS}{Multi-layer Analysis for Confidence Scoring}
\newacronym{ml}{ML}{Machine Learning}

\newacronym{ood}{OoD}{Out-of-Distribution}

\newacronym{pca}{PCA}{Principal Component Analysis}
\newacronym{pgd}{PGD}{Projected Gradient Descent}

\newacronym{sa}{SA}{Square Attack}
\newacronym{svd}{SVD}{Singular Value Decomposition}
\newacronym{sota}{SOTA}{State-Of-The-Art}
\newacronym{svm}{SVM}{Support Vector Machine}

\newacronym{vgg}{VGG}{Visual Geometry Group}

\newacronym{xai}{XAI}{eXplainable Artificial Intelligence}

%% file: tex/definitions.tex
\newcolumntype{M}[1]{>{\centering\arraybackslash}m{#1}}

\SetKwInOut{Input}{Input}
\SetKwInOut{Output}{Output}

\newcommand{\codeFont}[1]{\texttt{#1}\xspace}
\newcommand{\ray}{Ray Tune\xspace}
\newcommand{\optuna}{Optuna\xspace}

\newcommand{\realSet}{\mathbb{R}}
\newcommand{\tensor}[1]{\ensuremath{\bm{{#1}}}}
\newcommand{\norm}[1]{\ensuremath{\left\lVert#1\right\rVert}}
\newcommand{\mean}{\ensuremath{\mu}}
\newcommand{\cov}{\ensuremath{\Sigma}}
\newcommand{\setFont}[1]{\ensuremath{\mathcal{#1}}}
\newcommand{\grad}{\ensuremath{\nabla}}
\DeclareMathOperator*{\argmax}{\ensuremath{arg\,max}}

\newcommand{\ones}{\ensuremath{\tensor{1}}}
\newcommand*{\defeq}{\stackrel{\text{def}}{=}}
\newcommand{\auc}{\ensuremath{\Lambda}}
\newcommand{\aucGeom}{\overline{\auc}}
\newcommand{\aucGeomAA}{\aucGeom_{\rm{AA}}}
\newcommand{\aucGeomOOD}{\aucGeom_{\rm{OoD}}}
\newcommand{\aucGeomAll}{\aucGeom_{\rm{all}}}

\newcommand{\vgg}{VGG\xspace}
\newcommand{\mobilenet}{MobileNet\xspace}
\newcommand{\resnet}{Resnet\xspace}
\newcommand{\convnext}{ConvNeXt\xspace}
\newcommand{\layerName}[1]{\codeFont{#1}}

\newcommand{\cifar}{CIFAR-100\xspace}
\newcommand{\places}{Places365\xspace}
\newcommand{\SVHN}{SVHN\xspace}
\newcommand{\imagenet}{ImageNet1k\xspace}
\newcommand{\mnist}{MNIST\xspace}
\newcommand{\textures}{DTD\xspace}
\newcommand{\iNaturalist}{iNaturalist\xspace}
\newcommand{\openImageO}{OpenImage-O\xspace}

\newcommand{\dataset}{\setFont{D}}
\newcommand{\trainingSet}{\dataset^{\rm train}\xspace}
\newcommand{\validationSet}{\dataset^{\rm val}\xspace}
\newcommand{\testSet}{\dataset^{\rm test}\xspace}
\newcommand{\setSize}[1]{|#1|}
\newcommand{\trainingSetSize}{\setSize{\trainingSet}}
\newcommand{\validationSetSize}{\setSize{\validationSet}}
\newcommand{\testSetSize}{\setSize{\testSet}}
\newcommand{\numClasses}{\ensuremath{C}}

\newcommand{\nn}{\Phi}
\newcommand{\img}{\tensor{\xi}}
\newcommand{\lab}{\ensuremath{l}}
\newcommand{\prediction}{\hat{\lab}}
\newcommand{\numLayers}{\ensuremath{L}}

\newcommand{\layerIn}{\tensor{x}}
\newcommand{\layerOut}{\tensor{y}}
\newcommand{\weights}{\tensor{w}}
\newcommand{\bias}{\ensuremath{\tensor{b}}}
\newcommand{\kernel}{\ensuremath{\tensor{k}}}

\newcommand{\channel}{\ensuremath{c}}
\newcommand{\channelsIn}{\channel_{\rm{i}}}
\newcommand{\channelsOut}{\channel_{\rm{o}}}
\newcommand{\kernelWidth}{\ensuremath{w_{\rm{k}}}}
\newcommand{\kernelHeight}{\ensuremath{h_{\rm{k}}}}
\newcommand{\inputWidth}{\ensuremath{w_{\rm{i}}}}
\newcommand{\inputHeight}{\ensuremath{h_{\rm{i}}}}
\newcommand{\outputWidth}{\ensuremath{w_{\rm{o}}}}
\newcommand{\outputHeight}{\ensuremath{h_{\rm{o}}}}

\newcommand{\paddingWidth}{\ensuremath{w_{\rm p}}}
\newcommand{\paddingHeight}{\ensuremath{h_{\rm p}}}
\newcommand{\dilationWidth}{\ensuremath{w_{\rm d}}}
\newcommand{\dilationHeight}{\ensuremath{h_{\rm d}}}
\newcommand{\strideWidth}{\ensuremath{w_{\rm s}}}
\newcommand{\strideHeight}{\ensuremath{h_{\rm s}}}

\newcommand{\affineWB}{\tensor{B}}
\newcommand{\score}{\ensuremath{s}}
\newcommand{\SVDU}{\tensor{R}}
\newcommand{\SVDV}{\tensor{S}}
\newcommand{\SVDS}{\tensor{P}}
\newcommand{\approxLayerOut}{\tilde{\layerOut}}

\newcommand{\corevector}{corevector\xspace} 
\newcommand{\corevectors}{corevectors\xspace} 
\newcommand{\coreVec}{\tensor{\nu}}
\newcommand{\coreVecSize}{\kappa}

\newcommand{\peephole}{peephole\xspace} 
\newcommand{\peepholes}{peepholes\xspace} 
\newcommand{\ph}{\tensor{\rho}}

\newcommand{\coreVecAvg}{\coreVec^{\rm{a}}}
\newcommand{\coreVecToeplitz}{\coreVec^{\rm{t}}}
\newcommand{\coreVecKernel}{\coreVec^{\rm{k}}}
\newcommand{\toeplitzDimRed}{\codeFont{toeplitz}}
\newcommand{\avgpDimRed}{\codeFont{avgpooling}}
\newcommand{\kernelDimRed}{\codeFont{kernel}}

\newcommand{\DMDmean}{\tensor{\mean}^{\rm{DMD}}}
\newcommand{\DMDcov}{\tensor{\cov}^{\rm{DMD}}}
\newcommand{\DMDmagnitude}{\ensuremath{\epsilon}}
\newcommand{\DMDph}{\ph^{\rm{DMD}}}
\newcommand{\DMDdecisionFlow}{\ensuremath{\tensor{l}}}
\newcommand{\DMDlogisticRegressor}{\ensuremath{\Omega}}
\newcommand{\DMDscore}{\score^{\rm{DMD}}}

\newcommand{\MACScluster}{\ensuremath{\Theta}}
\newcommand{\MACSnumClusters}{\ensuremath{F}}
\newcommand{\MACSclusterProbs}{\ensuremath{\tensor{g}}}
\newcommand{\MACSempiricalPosterior}{\tensor{U}}
\newcommand{\MACSph}{\ph^{\rm{MACS}}}

\newcommand{\classificationmap}{classification-map\xspace} 
\newcommand{\classificationmaps}{classification-maps\xspace} 
\newcommand{\protoclasses}{proto-maps\xspace} 
\newcommand{\MACSprotoSet}{\setFont{T}}
\newcommand{\MACSclsmap}{\tensor{G}}
\newcommand{\MACSprotoc}{\ensuremath{\tensor{T}}}
\newcommand{\MACSth}{\ensuremath{\delta}}
\newcommand{\MACSscore}{\score^{\rm{MACS}}}

\newcommand{\svdKernel}{\hat{\kernel}}
\newcommand{\unfoldedLayerIn}{\hat{\layerIn}}
\newcommand{\isomorphWeights}{\hat{\weights}}
\newcommand{\isomorphAffineWB}{\hat{\affineWB}}
\newcommand{\isomorphSVDU}{\hat{\SVDU}}
\newcommand{\isomorphSVDV}{\hat{\SVDV}}
\newcommand{\isomorphSVDS}{\hat{\SVDS}}

%% file: sections/header.tex
\let\WriteBookmarks\relax
\def\floatpagepagefraction{1}
\def\textpagefraction{.001}

\shorttitle{Convolutional Layer Dimensionality Reduction for OoD and AA Detection Methods}    

\shortauthors{de Souza Rosa, L., Et al.}

\title [mode = title]{A Convolutional Layer Activation Dimensionality Reduction for Out-of-Distribution and Adversarial Attack Detection Methods}  

\author[1]{Leandro {de Souza Rosa}}[orcid=0000-0003-3457-9164]
\ead{leandro.desouzarosa@unibo.it}
\cormark[1]
\fnmark[1]
\credit{Conceptualization, Methodology, Software, Validation, Writing, Visualization}

\author[1]{Lorenzo Capelli}[orcid=0009-0005-4378-9982]
\ead{l.capelli@unibo.it}
\credit{Conceptualization, Software, Writing}

\author[1]{Clara Nunes Barrancos}[orcid=0009-0004-5713-5472]
\ead{clara.barrancos@unibo.it}
\credit{Conceptualization, Software, Writing}

\author[1,2]{Mauro Mangia}[orcid=0000-0002-3818-9115]
\ead{mauro.mangia@unibo.it}
\credit{Supervision, Project administration, Writing}

\author[1,2]{Riccardo Rovatti}[orcid=0000-0002-4731-7860]
\ead{riccardo.rovatti@unibo.it}
\credit{Supervision, Project administration, Writing}

\affiliation[1]{
	organization={Dipartimento di Ingegneria dell'Energia Elettrica e dell'Informazione ``Guglielmo Marconi'' (DEI) - Alma Mater Studiorum Università di Bologna},
	city={Bologna},
	country={Italy}
}

\affiliation[2]{
	organization={Advanced Research Center on Electronic Systems ``Ercole De Castro'' (ARCES) - Alma Mater Studiorum Università di Bologna},
	city={Bologna},
	country={Italy}
}

\cortext[1]{Corresponding author}



%% file: sections/abstract.tex
\begin{abstract}
	Despite the success of convolutional neural networks in image classification tasks and their general application in multi-modal models, their susceptibility to out-of-distribution and adversarial attack samples raises concerns regarding trustworthiness and safety.
    Among the approaches to tackle such issues, detection methods that analyze the model's intermediate activations to estimate a confidence score are a promising family that evaluates the decision process, relying on a dimensionality reduction step to enable efficient downstream processing of the high-dimensional activations.
    However, when considering convolutional layers, the dimensionality reduction methods in the literature either lack a mechanism to control the compression/information-loss trade-off or yield large representations.
	In this paper, we carefully analyze two state-of-the-art detection methods and their dimensionality reductions for convolutional layers and develop a novel reduction method with a controllable high-compression level.
	We extend these two state-of-the-art detection methods, enabling the usage of any dimensionality reduction, and evaluate their performance on out-of-distribution and adversarial attack detection.
	Results show that the detection methods with the proposed dimensionality reduction consistently perform better than, or comparable to, the strongest alternative.
	Furthermore, the proposed method is shown to reduce computation and memory footprints, given that it has the highest compression among the compared methods.
\end{abstract}


\begin{keywords}
	Convolutional Neural Networks \sep Dimensionality Reduction \sep Internal Activation Analysis \sep Out-of-Distribution Detection \sep Adversarial Attacks Detection
\end{keywords}

%% file: sections/introduction.tex
\section{Introduction}\label{sec: intro}

The widespread application of \glspl{dnn} today raises concerns about their safety and trustworthiness, problems that have recently been formalized in the \gls{ai} Act \cite{cancela2024eu}.
Among such models, well-known \glspl{cnn} are often used as image classifiers, or as part of multimodal models, e.g., \cite{radford2021learning,song2018deep,segura2022multimodal}, since they outperform transformer-based models, despite their susceptibility to \gls{ood} and \gls{aa} samples that can easily mislead their predictions.

While methods for enhancing the robustness of \glspl{dnn} have been proposed based on image manipulation \cite{xu2017feature}, or the model's outputs \cite{dadalto2023data}, solutions that analyze the model's \glspl{ia} estimate a confidence score based on the typicality of its decision process.
Among the latter, \gls{dmd} \cite{lee2018simple} and \gls{macs} \cite{capelli2025multi} analyze the \glspl{llf} within the model's \glspl{ia}, matching them to layer-wise human-understandable \glspl{hlf}, and provide a measurable metric about the model's internal decision process that is used for \gls{ood} and \gls{aa} detection by analyzing multiple layers.

When considering \glspl{cnn}, the sheer size of the model's \glspl{ia} \cite{pintelas2023multi} presents a major limitation for detection methods, especially when considering the ever-larger modern models, a problem countered by applying \gls{dr} methods, e.g., average-based, \gls{svd}, or \gls{ica} \cite{chen2018detecting}, which implies the necessity of balancing the compression level and information loss for efficient processing without hindering the downstream analysis.

Among such methods, this paper focuses on the well-established \gls{dmd} \cite{lee2018simple}, which performs a high-compression average-based \gls{dr} on \glspl{cl}' \glspl{ia}, and \gls{macs} \cite{capelli2025multi}, which applies an \gls{svd}-based \gls{dr}, allowing control over the compression level.
Particularly, we perform a deep analysis of \gls{dmd} and \gls{macs}, highlighting their mechanics and shortcomings to develop a novel \gls{dr} method for \glspl{cl} with high compression and controllable compression level.

The contributions of this paper are\footnote{Source code: \url{https://github.com/SSIGPRO/Convolutional-Layer-Reduction-for-Intermediate-Layer-Analysis}}:
\textbf{i.} A novel high-compression \gls{dr} method for \glspl{cl}' activations for \gls{ood} and \gls{aa} detection methods;
\textbf{ii.} Extensions of \gls{dmd} and \gls{macs} allowing them to work with generic \gls{dr} methods;
\textbf{iii.} An extensive evaluation showing that the proposed method maintains meaningful information despite its high compression, leading to comparable or higher detection performance.

The rest of this paper is organized as follows:
Section \ref{sec: related works} presents relevant related works.
Section \ref{sec: reference methods} presents \gls{dmd} and \gls{macs} in detail, highlighting their strengths, shortcomings, and mechanics.
Sections \ref{sec: method} and \ref{sec: results} present the proposed method and its evaluation, respectively.
Finally, Section \ref{sec: conclusions} concludes the paper.

%% file: sections/relatedWorks.tex
\section{Related Works}\label{sec: related works}

The myriad of \gls{ood} and \gls{aa} detection methods for image classifiers can be divided into two macro families:
``ante-hoc'' methods, which are \glspl{dnn} with embedded detection mechanisms, requiring specific training and hence not applicable to pre-existing models \cite{liu2020simple,fort2021exploring};
and ``post-hoc'' methods, which assume a pre-trained \gls{dnn}.
Among the latter, three main sub-families arise.



The first family consists of methods that manipulate the \glspl{dnn}'s inputs, being completely agnostic to the model itself, and generally work by creating sample variations through filtering \cite{xu2017feature,liang2021detecting} or perturbations \cite{jha2019attribution} and evaluating the model's predictions on them.
However, such methods are often limited in scope, e.g., to image classifiers, and are vulnerable to \glspl{aa} devised to avoid detection \cite{he2017adversarial}.


The second family consists of approaches that focus on the model's outputs.
Expanding on the fact that \glspl{cnn}' outputs are poorly calibrated \cite{minderer2021revisiting}, these approaches create lightweight calibration functions \cite{guo2017calibration, perez2022beyond} evaluating only the model's outputs.
Modern approaches in this family create approximations of Monte Carlo dropout models \cite{gal2016dropout} using statistical tests \cite{roth2019odds} or epistemic uncertainty estimation \cite{dadalto2023data,yoon2024uncertainty} using samples from a reference dataset, overcoming the distribution shift problems in calibration methods \cite{singh2021dark,wang2021rethinking,balanya2024adaptive}.

The third family consists of methods that evaluate the \glspl{dnn}' \glspl{ia}.
As examples, \cite{harder2021spectraldefense} uses Fourier transforms on the activations and a logistic regressor trained on nominal and \gls{aa} samples, and \cite{gorbett2022utilizing} proposes a \gls{svm} on the penultimate layer's activations to identify misclassifications, \gls{ood}, and \gls{aa} samples.
However, such methods lack a \gls{dr} step, limiting their scalability to larger models and motivating \cite{chen2018detecting} to use an \gls{ica}-based \gls{dr} and clustering to identify back-door \glspl{aa} based on the model's last hidden layers.
As such, methods seldom perform analysis based on the \glspl{ia} of earlier layers, especially for convolutional models, whose \glspl{cl} have extremely high dimensionality, thereby requiring non-scalable deep learning methods such as the \gls{knn}-based \cite{papernot2018deep} and the \gls{cam}-based \cite{rossolini2022increasing} approaches to produce a confidence score by comparing the activations of an inference sample w.r.t. the ones from a reference set.

Works that evaluate \glspl{dnn}' \glspl{ia} and leverage \gls{dr} methods are the focus of this paper.
\gls{dmd} \cite{lee2018simple} focuses specifically on \glspl{cl}, applying a channel-wise high-compression \gls{dr} to their activations, and estimating the prediction agreement throughout multiple layers with a logistic regressor trained on \gls{ood}/\gls{aa} samples.
Extensions of \gls{dmd} have been proposed by extending its feature association step with layer- and channel-wise \glspl{kde} to obtain class-agnostic representations that are classified with a logistic regressor \cite{erdil2021task}; and by enriching its dimensionality-reduced representation with a compression-based encoding \cite{yu2021convolutional}\footnote{\gls{dmd}'s base form, see Section \ref{subsec: dmd}.}.
\gls{macs} \cite{capelli2025multi} uses an \gls{svd}-based compression, overcoming \gls{dmd}'s restriction to \glspl{cl}, and an unsupervised clustering with an empirical posterior analysis to characterize the model's decision process throughout its layers over nominal samples, overcoming the need for \gls{ood}/\gls{aa} samples for \gls{dmd}'s regressor fitting.


Notably, reducing the dimensionality of \glspl{cl} is a practical approach, since their high-dimensional features are known for encoding low-dimensional \glspl{llf} within their manifold \cite{recanatesi2019dimensionality,ansuini2019intrinsic}.
%
Such results are often leveraged by model-compression methods to create compact versions of models without sacrificing their performance, being successfully applied to models ranging from \glspl{cnn}
\cite{meneghetti2023dimensionality} to \glspl{llm} \cite{sakr2024espace}.

%% file: sections/referenceMethods.tex
\section{Reference Methods}\label{sec: reference methods}

\subsection{Preliminaries}\label{subsec: preliminaries}

We target \gls{cnn} image classifiers ($\nn$), whose inputs are multi-channel images ($\img$) associated with labels ($\lab$) and predictions $\prediction=\nn(\img)$ belonging to $\numClasses$ classes. We consider datasets organized into training ($\trainingSet$), validation ($\validationSet$), and test ($\testSet$) splits\footnote{Notation wise, we use bold symbols for tensors, including vectors and matrices, and non-bold ones for scalar values.}.

\begin{figure}
	\centering
	\includegraphics[width=0.7\linewidth]{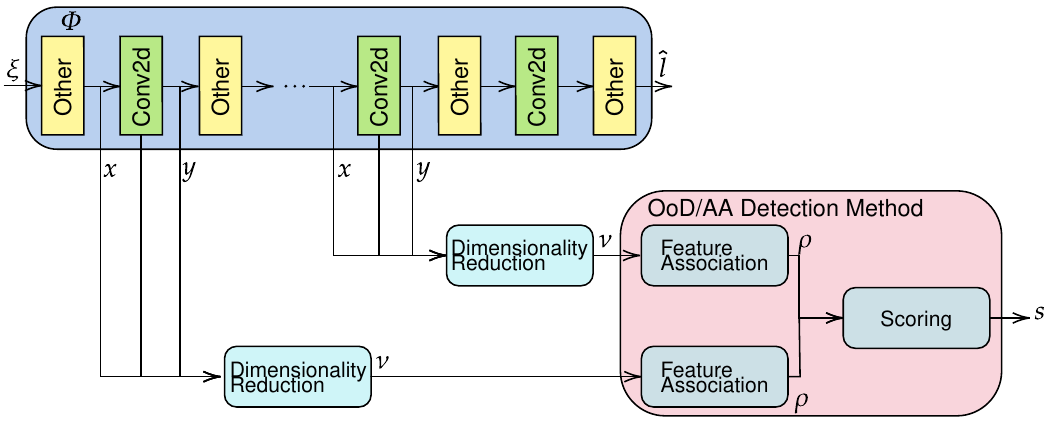}
	\caption{General overview of \glspl{ia}-based methods for \gls{ood} and \gls{aa} detection.}
	\label{fig: conf methods overview}
\end{figure}

We consider \gls{ood} and \gls{aa} \glspl{ia}-based detection frameworks that fit into the three-stage pipeline illustrated in Figure \ref{fig: conf methods overview}.
The first step consists of a \gls{dr} to obtain a compact version of the activations, referred to as ``\corevectors'' ($\coreVec$) hereinafter, which capture the \glspl{llf} present within the \glspl{ia} to enable efficient downstream processing.
In the second step, ``feature association'', the \corevectors are associated with \glspl{hlf}\footnote{Commonly, the labels $\lab$ are used as \glspl{hlf}.}, resulting in tensors that encode the features' presence at each layer, referred to as ``\peepholes'' ($\ph$) hereinafter.
The third step, ``scoring'', estimates a ``confidence score'' ($\score$) based on the \peepholes from a set of selected layers, estimating whether the sample is \gls{id}/nominal or \gls{ood}/\gls{aa}.

Note that approaches in the literature may implement all three steps, e.g., \gls{dmd} \cite{lee2018simple} and \gls{macs} \cite{capelli2025multi}, or skip some steps, e.g., \cite{rossolini2022increasing,harder2021spectraldefense,gorbett2022utilizing} do not leverage a \gls{dr} step.
Furthermore, it is common practice to attach the $\coreVec$'s and $\ph$'s computation to a subset of $\numLayers$ layers to limit the computational overhead.

Nevertheless, the \gls{dr} step becomes extremely important since the high-dimensional raw \glspl{ia} hinder most analysis in real scenarios, especially when the ever-increasing size of modern \glspl{dnn} is taken into consideration.
Therefore, \gls{dr} methods must also ensure that important \glspl{llf} present within the \glspl{ia} are captured in their result.
Special attention must be given to \glspl{cl}, whose activations span higher dimensions than their linear counterparts, making it difficult for \gls{dr} methods to effectively compress the activations without significant information loss.
Hence, \gls{dr} methods should also provide control over the trade-off between compression and information loss, enabling the user to tune the detection method.

Our goal is to create a \gls{dr} method for \glspl{cl} and evaluate its effect on \gls{dmd} and \gls{macs}; thus, we consider the detection methods to be composed of the feature association and scoring steps only, regardless of which compression method is used.
Particularly, we leverage lessons from \gls{dmd}'s and \gls{macs}'s original \glspl{dr} to achieve a controllable high-compression rate.

\subsection{Convolutional Layers}\label{subsec: conv layers}

\acrfullpl{cl} take as input tensors\footnote{Images are considered as tensors.} $\layerIn\in\realSet^{\channelsIn\times\inputHeight\times\inputWidth}$, resulting in output ones $\layerOut\in\realSet^{\channelsOut\times\outputHeight\times\outputWidth}$, where $(\channelsOut,\channelsIn)$ are the number of output and input channels, and ($\inputHeight, \inputWidth$) and ($\outputHeight, \outputWidth$) are the input's and output's height and width, respectively.
A \gls{cl} is characterized by its learned kernels $\kernel_{i,j}\in\realSet^{\kernelHeight\times\kernelWidth} ~|~ i \in \{0, \hdots, \channelsOut-1\}, j \in \{0, \hdots, \channelsIn-1\}$, where ($\kernelHeight$, $\kernelWidth$) are their height and width, the learned bias $\bias\in\realSet^{\channelsOut}$, and height and width parameters for their padding $(\paddingHeight, \paddingWidth)$, dilation $(\dilationHeight, \dilationWidth)$, and stride $(\strideHeight, \strideWidth)$.

Mathematically, the convolution operation is written as Eq. \eqref{eq: convolution}, where $\star$ is the $2$-D cross-correlation operator.

\begin{equation}\label{eq: convolution}
	\layerOut_{i} = \bias_{i}+\displaystyle\sum_{j=0}^{\channelsIn-1}\kernel_{j,i}\star\layerIn_j ~\forall~ i\in\{0, \hdots, \channelsOut-1\}
\end{equation}

The padding, dilation, and stride control how the kernels are applied during the cross-correlation, leading to the output sizes described in Eq. \eqref{eq: conv out shape}.

\begin{equation}\label{eq: conv out shape}
	\begin{aligned}
		\outputHeight = & \left\lfloor\frac{\inputHeight+2\paddingHeight-\dilationHeight(\kernelHeight-1)-1}{\strideHeight}+1\right\rfloor \\
		\outputWidth = & \left\lfloor\frac{\inputWidth+2\paddingWidth-\dilationWidth(\kernelWidth-1)-1}{\strideWidth}+1\right\rfloor
	\end{aligned}
\end{equation}

\subsection{DMD}\label{subsec: dmd}

\gls{dmd} \cite{lee2018simple} is a well-established method for \gls{ood} and \gls{aa} detection based on the Mahalanobis distance between artificial samples created from perturbations and a reference set.

\subsubsection{Average Pooling Dimensionality Reduction}\label{subsubsec: avgp dim red}

\gls{dmd}, as originally proposed, focuses on \glspl{cl}, leveraging the average pooling function described in Eq. \eqref{eq: avgp dim red} as its \gls{dr}, which yields \corevectors of a fixed dimension $\coreVecAvg\in\realSet^{\channelsOut}$.
We will refer to this \gls{dr} as \avgpDimRed hereinafter.

\begin{equation}\label{eq: avgp dim red}
	\coreVecAvg_{i} = \frac{1}{\outputHeight\outputWidth}
	\displaystyle
	\sum_{l=0}^{\outputHeight-1}
	\sum_{m=0}^{\outputWidth-1}
	\layerOut_{i, l, m}
	~\forall~i \in \{0, \hdots, \channelsOut-1\}
\end{equation}

At a high level, the idea behind \avgpDimRed is to capture the \glspl{llf}' presence/absence in each of the \gls{cl}'s output channels, disregarding their spatial localization within the activations.
As a consequence, \avgpDimRed can only be used on \glspl{cl}, limiting \gls{dmd}'s general applicability, although it can be applied to linear layers by skipping the \gls{dr} step \cite{capelli2025multi}.
The drawback is the lack of a mechanism to balance the compression/information-loss trade-off due to its fixed $\coreVecAvg$'s size, which can be problematic for modern \glspl{cnn} given their large number of channels.

\subsubsection{Feature Association}\label{subsubsec: dmd ph}

In this step, inference \corevectors are compared against the ones from a reference set\footnote{Typically, the training set.} using a \gls{hlf}-wise Mahalanobis distance.
First, a subset is created with samples of each class in the training set, i.e., $\trainingSet_i = \{\img_j \in \trainingSet~|~\lab_j=i\}$, and their respective $\coreVec$ average and overall covariance are computed according to Eq. \eqref{eq: dmd mean cov}\footnote{Note the notation agnostic to the \gls{dr} method used, written with the generic \corevectors $\coreVec$.}.

\begin{equation}\label{eq: dmd mean cov}
	\begin{aligned}
		\DMDmean_i = & \frac{1}{|\trainingSet_i|}\displaystyle\sum_{j=0}^{|\trainingSet_i|-1}\coreVec_j  ~\forall~ i \in \{0, \hdots, \numClasses-1\} \\
		\DMDcov = & \frac{1}{|\trainingSet|}\displaystyle\sum_{i=0}^{\numClasses-1}\sum_{j=0}^{|\trainingSet_i|-1}\left(\coreVec_j-\DMDmean_i\right)\left(\coreVec_j-\DMDmean_i\right)^\top
	\end{aligned}
\end{equation}

\glspl{hlf} are associated with \peepholes $\DMDph$ encoding the likelihood of samples belonging to the $\numClasses$ classes according to the Mahalanobis distance between a sample's $\coreVec$ and the class-wise average and covariance, by computing variations of $\img$ which are perturbed with magnitude $\DMDmagnitude$ towards each possible class, according to Eq. \eqref{eq: dmd perturbations}, where $\grad_{\img}$ are the $\nn$'s gradients when $\img$ is given as input.

\begin{equation}\label{eq: dmd perturbations}
	\tilde{\img_i} = \img + \DMDmagnitude \, {\rm{sign}}\left(\grad_{\img}\left(\coreVec-\DMDmean_i\right){\DMDcov}^{-1}\left(\coreVec-\DMDmean_i\right)^\top\right)
\end{equation}

The \peephole is then computed using $\tilde{\img_i}$, as described in Eq. \eqref{eq: dmd ph}, where $\tilde{\coreVec_i}$ is the \corevector w.r.t. the layer's activations when $\tilde{\img_i}$ is given as input.

\begin{equation}\label{eq: dmd ph}
	\DMDph_i = -\frac{1}{2}\left(\tilde{\coreVec_i}-\DMDmean_i\right){\DMDcov}^{-1}\left(\tilde{\coreVec_i}-\DMDmean_i\right)^\top
\end{equation}

Note that this step requires a back-propagation step to compute $\tilde{\img}$, which is computationally expensive.
Therefore, as per the authors' recommendation, the $\coreVec$ and $\ph$ computations are attached only to the last \glspl{cl} of each macro-block within the \gls{cnn} under analysis, a decision we follow in our experiments.

\subsubsection{Score}\label{subsubsec: dmd score}

\gls{dmd}'s score estimates how consistently the \glspl{hlf} are present across the \gls{dnn}'s layers by creating a vector $\DMDdecisionFlow\in\realSet^{\numLayers}$ as described in Eq. \eqref{eq: dmd decision vec}, where the left index in $\leftindex_i\DMDph$ indicates the $i^{\rm{th}}$ layer's $\DMDph$.
As such, each element in $\DMDdecisionFlow$ captures the strongest \gls{hlf} present in each respective layer.

\begin{equation}\label{eq: dmd decision vec}
	\DMDdecisionFlow_i = \argmax(\leftindex_i\DMDph) ~\forall~ i \in \{0, \hdots, \numLayers-1\}
 \end{equation}

Next, a logistic regressor ($\DMDlogisticRegressor$) is fitted over the $\DMDdecisionFlow$s from two different datasets, one containing nominal samples\footnote{Usually from the validation set.} ($\validationSet_{\rm{ok}}$), and the other containing \gls{ood}/\gls{aa} samples ($\validationSet_{\rm{ko}}$), depending on the application.
After fitting, the regressor is used during inference to estimate the confidence score as $\DMDscore=\DMDlogisticRegressor(\DMDdecisionFlow)$.

While \gls{dmd} yields high accuracies for \gls{ood} and \gls{aa} detection, its regressor necessitates \gls{ood}/\gls{aa} samples prior to its application, which severely limits its usage in real-world scenarios where one cannot create such samples a priori without knowledge about their distribution or which attacks are used.
A common workaround is to fit the regressor on a varied set of \gls{ood}/\gls{aa} samples, and perform inference with images from other datasets/\glspl{aa}, also known as ``\gls{dmd}-unaware'', which often yields significantly worse performance \cite{capelli2025multi}.
Given that our goal is to evaluate the impacts of our proposed \gls{dr} method, we only consider \gls{dmd} in its aware version.
 
\subsection{MACS}\label{subsec: macs}

\gls{macs} \cite{capelli2025multi} is a general framework that uses an \gls{svd}-projection \gls{dr} to provide controllable compression, and is applicable to layers that can be written as an affine transformation, such as linear ones, or \glspl{cl} after their kernels are unrolled with a Toeplitz operator \cite{praggastis2022svd}.
The feature association step is composed of a \gls{gmm} clustering and a \gls{llf}-to-\gls{hlf} association based on \cite{liu2020explaining}, creating vectors that estimate the presence of \gls{hlf} in each layer, similar to \gls{dmd}.
And finally, the score compares the information from all layers using empirical estimates.

\subsubsection{Toeplitz-SVD Dimensionality Reduction}\label{subsubsec: toeplitz dim red}

A \gls{cl} can be written as an affine transformation according to Equation \eqref{eq: layer as affine}, where $\weights\in\realSet^{(\channelsOut\times\outputHeight\times\outputWidth)\times(\channelsIn\times\inputHeight\times\inputWidth)}$ is the Toeplitz unrolling of the convolution kernels $\kernel$\footnote{$\bias$'s broadcast is omitted for simplicity.}.

\begin{equation}\label{eq: layer as affine}
	\approxLayerOut = \left[
		\begin{array}{cc}
			\weights & \bias
		\end{array}
	\right]
	\left[
		\begin{array}{c}
			\layerIn \\
			\ones
		\end{array}
	\right]
	\defeq \affineWB\left[
	\begin{array}{c}
		\layerIn \\
		\ones
	\end{array}
	\right]
\end{equation}

Next, a trimmed \gls{svd} is applied to decompose $\affineWB\in\realSet^{(\channelsOut\times\outputHeight\times\outputWidth)\times(\channelsIn\times\inputHeight\times\inputWidth+1)}$ into the $\SVDS\in\realSet^{(\channelsOut\times\outputHeight\times\outputWidth)\times\coreVecSize}$, $\SVDU\in\realSet^{\coreVecSize\times\coreVecSize}$, and $\SVDV\in\realSet^{\coreVecSize\times(\channelsIn\times\inputHeight\times\inputWidth+1)}$ matrices such that $\affineWB\approx\SVDS\SVDU\SVDV$, where $\coreVecSize\leq\min(\channelsIn\inputHeight\inputWidth+1,\channelsOut\outputHeight\outputWidth)$\footnote{For \codeFont{bias}-less layers, the $+1$ is dropped.} is a user-defined parameter.

The \corevectors $\coreVecToeplitz\in\realSet^\coreVecSize$ are computed by applying the \gls{svd}'s projection to the input activation as described in Eq. \eqref{eq: toeplitz dim red}.
This \gls{dr}, referred to as \toeplitzDimRed hereinafter, allows controlling the \corevector's size according to the \gls{svd}'s trimming parameter $\coreVecSize$, hence enabling control over the compression/information-loss trade-off while the \gls{svd} guarantees minimum information loss as $\coreVecSize$ decreases.

\begin{equation}\label{eq: toeplitz dim red}
	\coreVecToeplitz = \SVDV\left[
	\begin{array}{c}
		\layerIn \\
		\ones
	\end{array}
	\right]
\end{equation}

The \toeplitzDimRed's shortcoming lies in high-dimensional, highly sparse matrices generated by the Toeplitz unrolling, typically requiring large $\coreVecSize$s to capture sufficient information, but compromising downstream processing.

\subsubsection{Feature Association}\label{subsubsec: macs ph}

In this step, a \gls{gmm} ($\MACScluster$) is fit over the \corevectors from a reference set\footnote{Typically, the training set.}, introducing a user-defined hyper-parameter to control the number of Gaussian distributions\footnote{Commonly referred to as ``number of features''.} ($\MACSnumClusters$) used to fit the data.
A cluster prediction results in the cluster assignment $\MACScluster(\coreVec)\in\{0, \hdots, \MACSnumClusters-1\}$ and in an associated cluster assignment probabilities vector $\MACSclusterProbs\in\realSet^{\MACSnumClusters}$ encoding the likelihood of $\coreVec$ to belong to each cluster.

Next, an empirical feature association step based on \cite{liu2020explaining} is applied over $\trainingSet$ to associate \glspl{llf}, at $\coreVec$-space, to \glspl{hlf}, resulting in the ``empirical posterior'' matrix $\MACSempiricalPosterior\in\realSet^{\numClasses\times\MACSnumClusters}$ described in Eq. \eqref{eq: macs empp}, which counts the number of samples for each pair of predicted classification $\nn(\img)$ and cluster assignment $\MACScluster(\coreVec)$, followed by a normalization on the clustering direction.

\begin{equation}\label{eq: macs empp}
	\begin{aligned}
		\MACSempiricalPosterior_{l,c} = & \left|\left\{\nn(\img)=l \wedge \MACScluster(\coreVec)=c\right\}\right| && \forall~ \img \in \trainingSet\\
		\MACSempiricalPosterior_{l,c} = &  \frac{\MACSempiricalPosterior_{l,c}}{\displaystyle\sum_{j=0}^{\numClasses-1}\MACSempiricalPosterior_{j,c}} && \forall \begin{array}{l}
			l \in \{0, \hdots, \numClasses-1\}, \\
			c \in \{0, \hdots, \MACSnumClusters-1\}
		\end{array}
	\end{aligned}
\end{equation}

Then, \gls{macs}'s \peephole is given by $\MACSph=\MACSempiricalPosterior\coreVec$.

\gls{macs}'s shortcoming lies in the \glspl{gmm}'s distance-based clustering, whose performance deteriorates for high-dimensional data \cite{aggarwal2001surprising}.
Particularly, the fact that \toeplitzDimRed requires large $\coreVecSize$s for minimizing the information loss imposes a theoretical upper bound on the sizes of the layer and its inputs, above which the \gls{gmm}'s clustering capabilities start to degrade.

\subsubsection{Score}\label{subsubsec: macs score}

\gls{macs}'s scoring evaluates the \peepholes from all $\numLayers$ target layers, comparing them against typical values from the reference set; hence avoiding \gls{dmd}'s shortcoming of necessitating \gls{ood}/\gls{aa} samples during its regressor training.

First, a ``\classificationmap'' $\MACSclsmap\in\realSet^{\numClasses\times\numLayers}$ is created for each sample by concatenating the $\MACSph$s from multiple layers, as described in Eq. \eqref{eq: macs cls map}, where the left index in $\leftindex_j\MACSph$ indicates the $j^{\rm{th}}$ layer's \peephole.
As such, a \classificationmap captures the model's decision process throughout its layers at \gls{hlf} level.

\begin{equation}\label{eq: macs cls map}
	\MACSclsmap_{i,j} = \leftindex_j\MACSph_i
	~\forall~\begin{array}{ll}
		i& \in \{0, \hdots, \numClasses-1\}, \\
		j& \in \{0, \hdots, \numLayers-1\}
	\end{array}
\end{equation}

Second, a set $\MACSprotoSet$ is created with $\MACSclsmap$s from $\trainingSet$ samples whose classification has high-confidence according to the model, i.e., $\MACSprotoSet_i=\{\MACSclsmap ~|~ \nn(\img)=i \land \max(\nn_{\rm{head}}(\img))>\MACSth ~\forall~ \MACSclsmap\in\trainingSet\}$, where $\MACSth$ is a user-defined threshold for the model's confidence, and the ``$\rm{head}$'' sub-script indicates $\nn$'s classification head.
As such, \classificationmaps within a $\MACSprotoSet_i$ represent the typical decision process for each class, which are summarized by a summation and normalization on the class direction to create the so-called ``\protoclasses'' $\MACSprotoc_i\in\realSet^{\numClasses\times\numLayers}$ according to Eq. \ref{eq: macs proto class}.

\begin{equation}\label{eq: macs proto class}
	\begin{aligned}
		\MACSprotoc_i = & \displaystyle\sum_{\MACSclsmap\in\MACSprotoSet_i}\MACSclsmap && \forall~ i ~\in~ \{0, \hdots, \numClasses-1\} \\
		[\MACSprotoc_i]_{c,l} = &  \frac{[\MACSprotoc_i]_{c,l}}{\displaystyle\sum_{j=0}^{\numClasses-1}[\MACSprotoc_i]_{j,l}} && \forall \begin{array}{ll}
			c & \in \{0, \hdots, \numClasses-1\}, \\
			l & \in \{0, \hdots, \numLayers-1\}
		\end{array}
	\end{aligned}
\end{equation}

Finally, a cosine-similarity is used to compute the score ($\MACSscore$) for inference samples based on their $\MACSclsmap$ and the predicted class's $\MACSprotoc$, according to Eq. \ref{eq: macs score}, where $\odot$ is the dot-product.

\begin{equation}\label{eq: macs score}
	\MACSscore = \frac{\MACSclsmap\odot\MACSprotoc_{\nn(\img)}}{\norm{\MACSclsmap}\norm{\MACSprotoc_{\nn(\img)}}}
\end{equation}

A major difference between \gls{macs}'s and \gls{dmd}'s scores is that the latter considers only the most prominent class in each layer (Equation \eqref{eq: dmd decision vec}), while the former considers all classes (Equation \eqref{eq: macs proto class}), capturing the fact that early layers typically present features from a subset of classes, instead of a single predominant one.

%% file: sections/method.tex
\section{Proposed Method}\label{sec: method}

\subsection{Overview}\label{subsec: proposed method overview}

\begin{figure}
	\centering
	\includegraphics[width=0.75\linewidth]{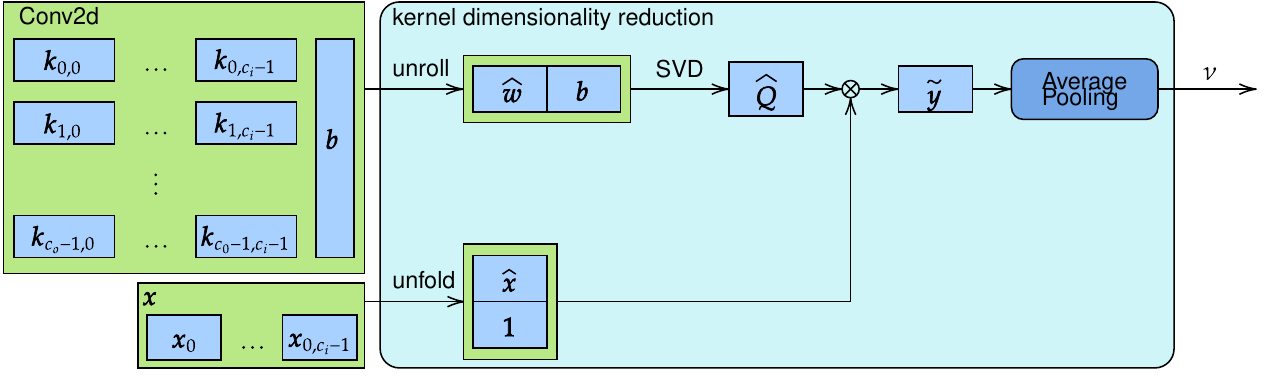}
	\caption{Proposed method \gls{cl}'s activation \gls{dr} overview.}
	\label{fig: proposed method overview}
\end{figure}

Figure \ref{fig: proposed method overview} presents an overview of the proposed method, which takes inspiration from \toeplitzDimRed and \avgpDimRed to achieve a highly compact representation of \glspl{cl}'s activations while allowing control over the compression/information-loss trade-off.
The main idea is to unroll the \gls{cl}'s kernels using the dense and compact convolution isomorphism explored in \cite{praggastis2022svd}, instead of the Toeplitz one used by \gls{macs}.
From this unrolling, we apply the \gls{svd} to create new kernels intrinsically ordered by importance ($\svdKernel$) and to enable control over the new number of output channels according to the \gls{svd}'s trimming parameter ($\coreVecSize$).
Finally, after applying the convolution to $\layerIn$ with this new kernel, we leverage \avgpDimRed, yielding a vector with the same dimension as $\coreVecSize$.
Therefore, the proposed method, referred to as \kernelDimRed hereinafter, yields a higher compression level than \avgpDimRed, with compression-level control through $\coreVecSize$, while guaranteeing minimal information loss.

\subsection{Detailed Description}\label{subsec: description}

\subsubsection{Convolution Isomorphism}\label{subsec: kernel unrolling}

The study presented in \cite{praggastis2022svd} shows that the Toeplitz unrolling is an isomorphism of the convolution operation, and its linearization in Eq. \eqref{eq: layer as affine} leads to a tensor with shape $\affineWB_{(\channelsOut\outputHeight\outputWidth), (\channelsIn\inputHeight\inputWidth+1)}$, and the layer's input must be reshaped into $\layerIn_{(\channelsIn\inputHeight\inputWidth+1), 1}$\footnote{Notably, this is the common \textit{flatten} operation.}.
It is important to note the discrepancy between the $\kernel$'s dimensions, which grow linearly with $(\channelsIn, \channelsOut, \kernelHeight, \kernelWidth)$ and their respective Toeplitz unrolling, which grows quadratically with the layer's input size $(\inputHeight, \inputWidth)$\footnote{Apply Eq. \eqref{eq: conv out shape} to $\affineWB$'s shape}.
%

An alternative isomorphism is studied in \cite{praggastis2022svd}\footnote{Denoted by ``W1'' in the paper.}, which simply reshapes the kernels to create $\isomorphWeights \in \realSet^{\channelsOut\times(\channelsIn\times\kernelHeight\times\kernelWidth)}$ by flattening and concatenating all input kernels for each output channel, as described in Eq. \eqref{eq: kernel reshape}.
Note that each row in $\isomorphWeights$ corresponds to an output channel and the columns to all respective input channels.

\begin{equation}\label{eq: kernel reshape}
	\isomorphWeights_{i, j\kernelHeight\kernelWidth+m\kernelWidth+n} = \kernel_{i, j, m, n}
	~\forall~ \begin{array}{l}
		i \in \{0, \hdots, \channelsOut-1\} \\
		j \in \{0, \hdots, \channelsIn-1\} \\
		m \in \{0, \hdots, \kernelHeight-1\} \\
		n \in \{0, \hdots, \kernelWidth-1\}
	\end{array}
\end{equation}

As a consequence, the input $\layerIn$ needs to be unfolded into $\unfoldedLayerIn\in\realSet^{(\channelsIn\times\kernelHeight\times\kernelWidth)\times(\outputHeight\times\outputWidth)}$ according to Algorithm \ref{alg: unfold}, where the element for each channel, row, and column of $\unfoldedLayerIn$ (lines $1$, $10$, and $12$) is mapped from $\layerIn$'s respective channel, row, and column (lines $4$, $6$, and $8$).
Note that the resulting tensor's dimensions are given by the layer's output ($\layerOut$), hence eliminating the quadratic scaling with the layer's input ($\layerIn$) dimension present in \toeplitzDimRed.

\begin{algorithm}[h]
	\caption{\gls{cl}'s input unfolding. ``\%'' and ``//'' are the reminder and integer division operators, respectively. Adapted from \cite{ansel2024pytorch}.}\label{alg: unfold}
	\Input{Layer's parameters $\kernel, \bias, (\channelsOut, \channelsIn), (\paddingHeight, \paddingWidth), (\strideHeight, \strideWidth), (\dilationHeight, \dilationWidth)$, input and output shapes $(\inputHeight, \inputWidth), (\outputHeight, \outputWidth)$, input $\layerIn$.}
	\Output{Unfolded input $\unfoldedLayerIn$.}
	\For{${\rm{channel}}_{\text{o}}\in\{0, \hdots, \channelsIn\kernelHeight\kernelWidth-1$\}}{
		${\rm{offset}}_{\text{w}} \leftarrow {\rm{channel}}_{\text{o}}\%\kernelWidth$\;
		${\rm{offset}}_{\text{h}} \leftarrow ({\rm{channel}}_{\text{o}}//\kernelHeight)\%\kernelHeight$\;
		${\rm{channel}}_{\text{i}} \leftarrow {\rm{channel}}_{\text{o}}//\kernelHeight//\kernelWidth$\;
		
		\For{${\rm{row}}_{\text{o}}\in\{0,\hdots,\outputHeight-1\}$}{
			${\rm{row}}_{\text{i}} \leftarrow {\rm{row}}_{\text{o}}\strideHeight-\paddingHeight+{\rm{offset}}_{\text{h}}\dilationHeight$\;
			\For{${\rm{col}}_{\text{o}}\in\{0,\hdots,\outputWidth-1\}$}{
				${\rm{col}}_{\text{i}} \leftarrow {\rm{col}}_{\text{o}}\strideWidth-\paddingWidth+{\rm{offset}}_{\text{w}}\dilationWidth$\;
				\uIf{${\rm{row}}_{\text{i}}\geq0 \land {\rm{col}}_{\text{i}}\geq0 \land {\rm{row}}_{\text{i}}<\inputHeight \land {\rm{col}}_{\text{i}} <\inputWidth $}{
					$\unfoldedLayerIn_{{\rm{channel}}_{\text{o}}, {\rm{row}}_{\text{o}}\outputWidth+{\rm{col}}_{\text{o}}} \leftarrow \layerIn_{{\rm{channel}}_{\text{i}},{\rm{row}}_{\text{i}},{\rm{col}}_{\text{i}}}$\;
				}
				\Else{
					$\unfoldedLayerIn_{{\rm{channel}}_{\text{o}}, {\rm{row}}_{\text{o}}\outputWidth+{\rm{col}}_{\text{o}}} \leftarrow 0$
				}
			}
		}
		
	}
\end{algorithm}

Next, the layer can be written in its affine form as described in Eq. \eqref{eq: isomorphism layer as affine} and the resulting $\isomorphAffineWB$ is a dense matrix with size depending only on the number of learned parameters within the \gls{cl}.

\begin{equation}\label{eq: isomorphism layer as affine}
	\approxLayerOut = \left[
	\begin{array}{cc}
		\isomorphWeights & \bias
	\end{array}
	\right]
	\left[
	\begin{array}{c}
		\unfoldedLayerIn \\
		\ones
	\end{array}
	\right]
	\defeq \isomorphAffineWB\left[
	\begin{array}{c}
		\unfoldedLayerIn \\
		\ones
	\end{array}
	\right]
\end{equation}

\subsubsection{Ordered Kernels Through SVD}\label{subsec: kernel svd}

We leverage the fact that each row in $\isomorphAffineWB$ corresponds to the kernels of each output channel.
The main idea is to decompose $\isomorphAffineWB$ with \gls{svd} into $\isomorphSVDS\in\realSet^{\channelsOut\times\coreVecSize}$, $\isomorphSVDU\in\realSet^{\coreVecSize\times\coreVecSize}$, and $\isomorphSVDV\in\realSet^{\coreVecSize\times(\channelsIn\times\kernelHeight\times\kernelWidth+1)}$ matrices such that $\isomorphAffineWB\approx\isomorphSVDS\isomorphSVDU\isomorphSVDV$, where $\coreVecSize\leq\min(\channelsIn\kernelHeight\kernelWidth+1, \channelsOut)$\footnote{For \codeFont{bias}-less layers, the $+1$ is dropped.}. We then use $\isomorphSVDV$'s projection to reduce $\layerIn$'s dimensionality, similarly to \gls{macs}.

To further compress the activations projected with $\isomorphSVDV$, we apply the average pooling (Eq. \eqref{eq: avgp dim red}), resulting in the proposed \corevector $\coreVecKernel\in\realSet^{\coreVecSize}$ described by Eq. \eqref{eq: kernel dim red}.
Note that $\coreVecSize\leq \min(\channelsIn\kernelHeight\kernelWidth+1, \channelsOut)$ ensures that the minimal compression is the same as \avgpDimRed.

\begin{equation}\label{eq: kernel dim red}
	\coreVecKernel_{i} = \frac{1}{\outputHeight\outputWidth}
	\displaystyle
	\sum_{l=0}^{\outputHeight-1}
	\sum_{m=0}^{\outputWidth-1}
	\left[\isomorphSVDV
	\left[\begin{array}{c}
		\unfoldedLayerIn \\
		\ones
	\end{array}\right]\right]_{i, l, m}
	~\forall~i \in \{0, \hdots, \coreVecSize-1\}
\end{equation}

It is important to notice that each row in $\isomorphSVDV$ corresponds to a new set of kernels, computed from $\isomorphAffineWB$'s projection into the eigenvector space, conveniently ordered according to the eigenvalues, such that reducing $\coreVecSize$ minimizes the information loss.
Therefore, using $\isomorphSVDV$ has the same advantages as \toeplitzDimRed.

\subsection{Implications for OoD and AA Methods}\label{subsec: implications}

Considering that we extend \gls{macs} and \gls{dmd} to accept different \gls{dr} methods, some considerations must be made.
Regarding \gls{dmd}, the \corevectors used for generating image variations through back-propagation should not be hindered by the \gls{dr}'s compression level, and the amount of information retained after compression should be the major factor in its performance; hence the method is expected to yield good performance even when using \toeplitzDimRed, whose typical $\coreVecSize$ is large.
Regarding \gls{macs}, its clustering step is expected to deteriorate for high-dimensional \corevectors; thus, one can expect better performance using \avgpDimRed and \kernelDimRed's compression if they successfully maintain the activations' information.

%% file: sections/results.tex
\section{Results}\label{sec: results}

\subsection{Experimental Setup}\label{subsec: experimental setup}

\subsubsection{Datasets}\label{subsubsec: datasets}

To evaluate the \kernelDimRed's impact on \gls{ood} and \gls{aa} detection methods, we select the well-known \cifar \cite{krizhevsky2009learning} as the in-distribution/nominal dataset, containing images of $\numClasses=100$ classes.
\gls{ood} samples are taken from \SVHN \cite{netzer2011reading}, \places \cite{zhou2017places}, \mnist \cite{lecun2002gradient}, and \textures \cite{cimpoi2014describing} from the ``Far-\gls{ood}'' standardized detection test for \cifar \cite{zhang2023openood}.

For testing the detection methods on a larger scale, we use \imagenet \cite{russakovsky2015imagenet} as the in-distribution/nominal dataset, containing samples from $1000$ classes.
In this case, \gls{ood} samples are taken from \iNaturalist \cite{van2018inaturalist}, \textures \cite{cimpoi2014describing}, and \openImageO \cite{wang2022vim} from the ``Far-\gls{ood}'' standardized detection test for \imagenet \cite{zhang2023openood}.

To generate \glspl{aa} samples for both \cifar and \imagenet, we apply the standard \gls{bim} \cite{BIM}, \gls{pgd} \cite{PGD}, plus the standardized detection tests \cite{croce2020robustbench} composed of \gls{fab} \cite{croce2020minimally}, \gls{sa} \cite{andriushchenko2020square}, \gls{apgd}, and \gls{apgd} using the cross-entropy and TRADES losses \cite{croce2020reliable}, to their $\validationSet$ and $\testSet$.

We randomly select a subset of images for each dataset, for \cifar $\trainingSetSize=20000$, for \imagenet $\trainingSetSize=50000$, and for all datasets $\validationSetSize=\testSetSize=5000$ samples except \textures, from which we take all $\setSize{\validationSet_{\textures}}=\setSize{\testSet_{\textures}}=1880$ samples.
Similarly, for \gls{dmd}, we use the $\validationSet_{\rm{ko}}=\validationSet$ corresponding to the same dataset $\testSet$ as it is tested (aware mode) since we are not interested in evaluating the method's robustness in its more realistic unaware mode.

\subsubsection{Models}\label{subsubsec: models}

Our tests consider $4$ standard \gls{cnn} image classifiers from \codeFont{TorchVision} \cite{marcel2010torchvision}:
\vgg \cite{simonyan2014very}, a traditional, well-studied architecture;
\mobilenet \cite{sandler2018mobilenetv2}, a lightweight model characterized by bottleneck inverted residual blocks and depth-wise convolution layers;
\resnet \cite{he2016deep}, characterized by residual convolutional blocks;
and \convnext \cite{liu2022convnet}, a modern state-of-the-art model that adopts depth-wise convolution layers and residual blocks, along with updates to activation and normalization functions.

We set \gls{dmd}'s and \gls{macs}'s activation processing (Figure \ref{fig: conf methods overview}) at the last \gls{cl} in the models' macro-blocks, according to \gls{dmd}'s guidelines:
For \vgg, \resnet, and \convnext, we use the \gls{cl} after each of their $4$, $4$, $3$ macro-blocks, respectively;
for \mobilenet, we empirically select $4$ macro-blocks maximizing the detection methods' \gls{auc} for \gls{ood} and \gls{aa} detection.
Table \ref{tab: layer selection} presents the selected layers and their parameters.

Interestingly, successive layers within the models have increasing $(\channelsIn, \channelsOut)$, but smaller $(\inputHeight, \inputWidth)$, leading the \glspl{llf} to be encoded channel-wise, motivating \gls{dr} methods such as \avgpDimRed and the proposed \kernelDimRed to consider output channels as a whole.

Note that this selection is fixed to showcase the \gls{dr}'s impact on the detection methods, leaving the exploration of more layers or different layer combinations outside the scope of this paper.

As a remark, the \codeFont{TorchVision} models, pre-trained over \imagenet ($1000$ classes), are fine-tuned on \cifar's $\trainingSet$ by substituting their heads ($\nn_{\rm{head}}$) to match $100$ classes and performing a standard training loop. The accuracies over \cifar's and \imagenet's $\testSet$ are reported in Table \ref{tab: layer selection}.

\input{tables/selectedLayers}

\subsubsection{Evaluation Metric}\label{subsubsec: eval metric}

The performance of \gls{dmd} and \gls{macs} is measured through the \gls{auc} ($\auc$) for differentiating between \gls{id}/nominal (from \cifar's or \imagenet's $\testSet$) and \gls{ood}/\gls{aa} samples\footnote{Samples are balanced. For \glspl{aa} we only take samples that would be correctly classified and had their prediction changed by the attacks.} based on their confidence score.
To summarize the results from multiple datasets, we use the geometric mean as a metric that severely punishes inconsistencies between different datasets, thus favoring robustness, when considering the \gls{ood} ($\aucGeomOOD$), \gls{aa} ($\aucGeomAA$), or all ($\aucGeomAll$) datasets, as described in Eq. \eqref{eq: auc geomeans}.

\begin{equation}\label{eq: auc geomeans}
	\begin{array}{lclll}
		\aucGeom & = & \sqrt[\setSize{\Delta}]{\displaystyle\prod_{i\in \Delta}\auc_i} & \Delta=\text{\gls{ood} and/or \gls{aa} datasets}\\
	\end{array}
\end{equation}

\subsection{Tuning}\label{subsec: tuning}

As highlighted in Sections \ref{sec: reference methods} and \ref{sec: method}, \gls{dmd}, \gls{macs}, \toeplitzDimRed, \avgpDimRed, and \kernelDimRed have hyper-parameters to be tuned for each layer, which are summarized in Table \ref{tab: hyperp}.
The threshold for \gls{macs} is fixed to $\MACSth=0.9$ for \cifar and $\MACSth=0$ for \imagenet.

\begin{table}[h!]
	\centering
	\caption{Hyper-parameters of \gls{dr} detection methods, and their lower (LB) and upper (UB) bounds.}
	\label{tab: hyperp}
	\begin{tabulary}{\linewidth}{l|C|C|C|C}
		\toprule
		Method & \toeplitzDimRed & \kernelDimRed & \gls{dmd} & \gls{macs} \\
		\midrule
		Param. & $\coreVecSize$ & $\coreVecSize$ & $\DMDmagnitude$ & $\MACSnumClusters$ \\\hline
		LB & $50$ & $50$ & $0.001$ & $50$\\\hline
		UB & $1024$ & $\min(\channelsOut, $ $\channelsIn\kernelHeight\kernelWidth+1)$ & $0.01$ & $5\numClasses$\\
		\bottomrule
	\end{tabulary}
\end{table}

We leverage the state-of-the-art tuning library \ray \cite{liaw2018tune} with \optuna \cite{optuna2019} as a search engine and the Asynchronous Hyper-Band \cite{li2018massively} scheduler to optimize the hyper-parameters for each layer, given all possible combinations of \gls{dr} and detection methods.
Each parameter range\footnote{Consistent with values presented in \cite{capelli2025multi}.} is divided into $10$ intervals, and \ray iteratively selects promising configurations to maximize $\aucGeomAll$, evaluating a total of $50$ combinations out of the $10^{\numLayers}$ possible ones for \avgpDimRed, and the $100^\numLayers$ possible ones for \toeplitzDimRed and \kernelDimRed.

\begin{figure*}[h!]
	\centering
	\begin{subfigure}{\linewidth}
		\centering
		\includegraphics[width=\linewidth]{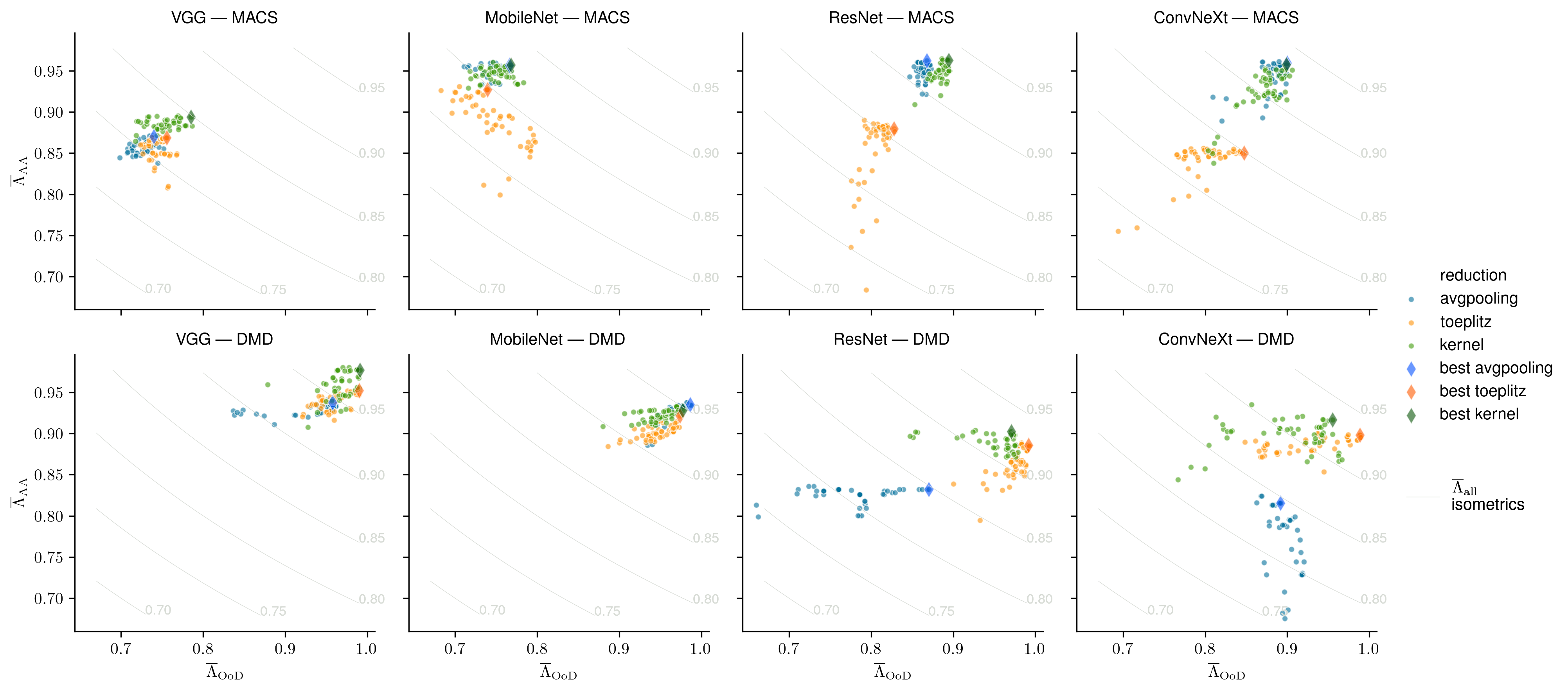}
		\caption{\cifar}
		\label{fig: tuning aucs cifar}
	\end{subfigure}
	\begin{subfigure}{\linewidth}
		\centering
		\includegraphics[width=\linewidth]{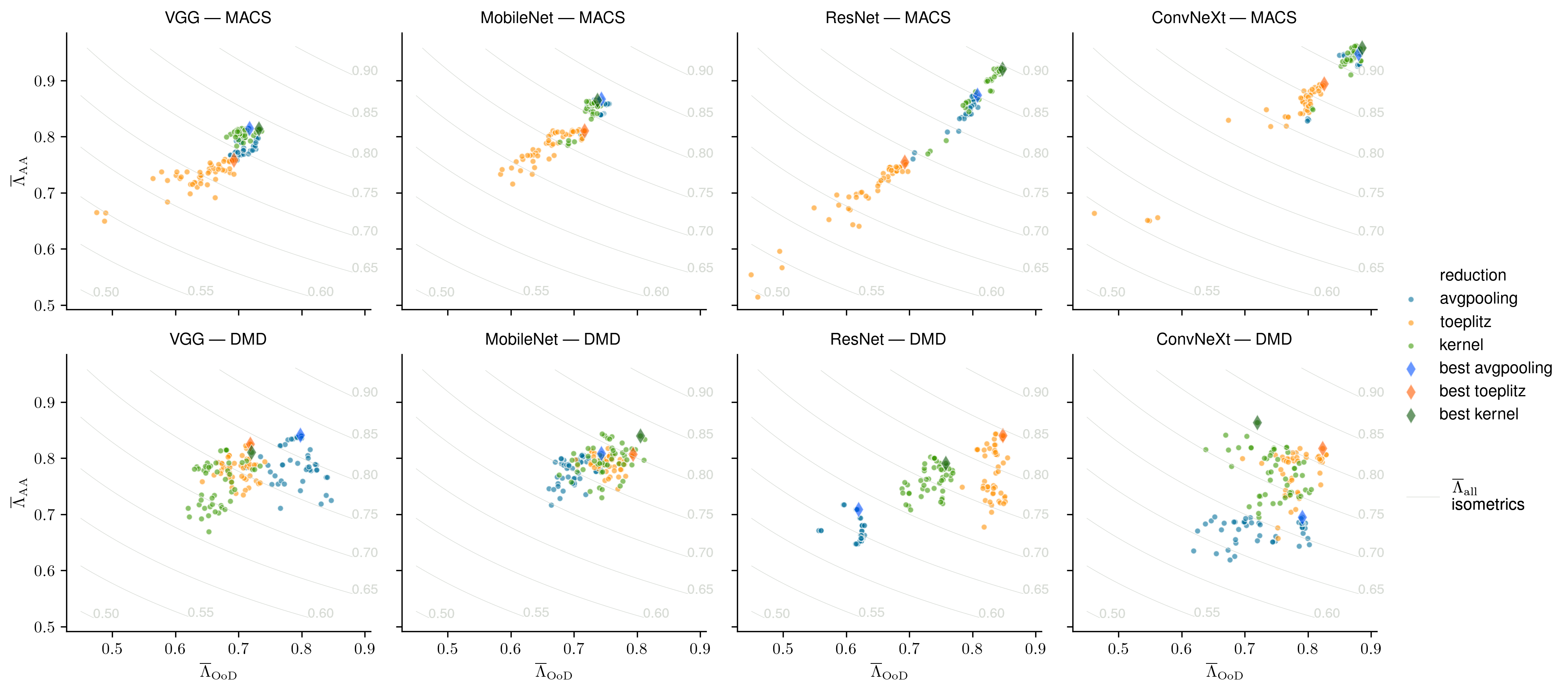}
		\caption{\imagenet}
		\label{fig: tuning aucs imagenet}
	\end{subfigure}
	\caption{$\aucGeomOOD$ and $\aucGeomAA$ for the configurations explored during tuning. The ``best'' ones, according to the maximum $\aucGeomAll$ (highlighted with a diamond-shaped mark) are chosen for further evaluation. Gray lines are the $\aucGeomAll$ product isometric curves.}
	\label{fig: aucs}
\end{figure*}

Figures \ref{fig: tuning aucs cifar} and \ref{fig: tuning aucs imagenet} present the $\aucGeomOOD$ and $\aucGeomAA$ for each combination of \gls{dr} and detection methods for \cifar and \imagenet, respectively.
\gls{macs} with \kernelDimRed matches or outperforms the other \glspl{dr} in all cases, and \gls{dmd} follows the same trend for \cifar, and for \imagenet with \mobilenet and \convnext, showing that \kernelDimRed successfully captures the information in each channel.

An interesting observation is that \gls{macs} typically performs better with \avgpDimRed w.r.t. its original \toeplitzDimRed, which is a direct consequence of the large $\coreVecSize$s hindering its \gls{gmm}.
On the other hand, \gls{dmd} generally performs better with \toeplitzDimRed w.r.t. its original \avgpDimRed, as expected since the \corevectors' size does not hinder its back-propagation step.
We highlight that allowing the detection methods to work with generic \glspl{dr} and showing these cross-results is one of our contributions.

As a side result, the larger spread of points in Figure \ref{fig: aucs} indicates a higher sensitivity to hyper-parameters, making the corresponding method less robust and more difficult to tune.
\gls{macs}'s sensitivity when using \toeplitzDimRed can be explained by the large $\coreVecToeplitz$ range, a consequence of its large sparse matrices, and \gls{dmd}'s sensitivity seems not to be affected by the \corevectors' dimensionality, despite results showing that \gls{dmd} is more sensitive to hyper-parameters than \gls{macs} in general.

Furthermore, results indicate a slightly stronger correlation between the \gls{ood} and \gls{aa} \glspl{auc} for \gls{macs}.
Given that \gls{macs} only needs $\trainingSet$, one tuning round makes the method robust for both \gls{ood} and \gls{aa} detection, while \gls{dmd}'s weaker correlation indicates that a tuning must be done for each application, which is an undesirable characteristic from the practical point of view.

Table \ref{tab: corevec size sum} summarizes the sum of $\coreVecSize$ for each \gls{dr}'s tuning configuration with the highest $\aucGeomAll$, showing how much more compact the proposed \kernelDimRed is than its upper bound given by \toeplitzDimRed.
Therefore, \kernelDimRed is expected to improve both memory and computation footprints, while achieving detection performance comparable to or higher than the best-performing alternative.

\input{tables/coreVecSizeRange}

\subsection{Memory and Computation Footprint}\label{subsec: computation footprint}

Table \ref{tab: mem and comp time} shows the memory used by the \gls{svd} matrices considering the largest value of $\coreVecSize$ from Table \ref{tab: hyperp}, showing that the proposed \kernelDimRed requires much less memory than \toeplitzDimRed.
Interestingly, for the residual architecture \resnet, the high $\channelsOut$ leads to large \avgpDimRed matrices, almost reaching the memory footprint of \toeplitzDimRed, displaying a case in which \kernelDimRed's maximum size is mostly beneficial.
Even though this step is done only once per model, we highlight that the \gls{svd} sizes directly impact the number of computations during the creation of the \corevectors and \peepholes.
In fact, \toeplitzDimRed scales poorly to larger models, e.g., its \gls{svd} computation overflows the GPU\footnote{NVIDIA A100 80GB RAM.} memory in our tests\footnote{The \gls{svd} was computed on CPUs.}, making it detrimental to larger modern models.

\input{tables/footprint}

Table \ref{tab: mem and comp time} also shows the memory and computation time\footnote{All computation times exclude data transfers.} for the \corevectors using each method.
While the proposed \kernelDimRed performs more steps (\gls{svd} projection and average pooling) than \avgpDimRed (average pooling only) and \toeplitzDimRed (projection only), the former's computational overhead is generally offset by its smaller $\coreVecSize$s, reducing the overall footprint.
Note that computation times for \gls{macs} and \gls{dmd} depend on other factors, e.g., \gls{macs}'s number of clusters, or \gls{dmd}'s GPU parallelization and memory access, making the relation between $\coreVecSize$ (Table \ref{tab: corevec size sum}) and computation time not completely straightforward.

Table \ref{tab: mem and comp time} also highlights that \gls{macs} is one order of magnitude faster than \gls{dmd}, being consistent with the results presented in \cite{capelli2025multi}.
Furthermore, we observe that \gls{dmd}'s timing increases more drastically for \imagenet w.r.t. \gls{macs}, which is explained by its number of inferences for each class (\peepholes computations), which are more expensive than the latter's per-class computations (score only).
Therefore, we highlight that another benefit of \gls{macs} is its scalability.

%% file: tables/selectedLayers.tex
\begin{table*}
	\centering
	\caption{Selected layers' details. The layer names are abbreviated according to: \layerName{f}, \layerName{c}, and \layerName{l} for ``features'', ``conv'', and ``layer'', respectively. All layers have $(\dilationHeight,\dilationWidth)=(1, 1)$.}
	\label{tab: layer selection}
	\begin{tabulary}{\linewidth}{L|L|L|C|C|C|C|C|C|C}
		\toprule
		\multirow{2}{*}{Model} & \multirow{2}{*}{Version} & \multirow{2}{*}{Layer Name} & \multirow{2}{*}{$(\kernelHeight, \kernelWidth)$} & \multirow{2}{*}{$(\channelsIn, \channelsOut)$} & \multirow{2}{*}{$(\paddingHeight, \paddingWidth)$} & \multirow{2}{*}{$(\strideHeight, \strideWidth)$} & \multirow{2}{*}{$(\inputHeight, \inputWidth)$} & \multicolumn{2}{c}{Acc. (\%)} \\\cline{9-10}
		& & & & & & & & \cifar & \imagenet \\\midrule
		\multirow{4}{*}{\vgg} & \multirow{4}{*}{\codeFont{vgg16}} & \layerName{f.7} & \multirow{4}{*}{$(3, 3)$} & $(128, 128)$ & \multirow{4}{*}{$(1, 1)$}  & \multirow{4}{*}{$(1, 1)$} & $(112, 112)$ & \multirow{4}{*}{$0.76$} & \multirow{4}{*}{$0.70$} \\
		&& \layerName{f.14} &  & $(256, 256)$ & & & $(56, 56)$ & & \\
		&& \layerName{f.21} &  & $(512, 512)$ & & & $(28, 28)$ & & \\
		&& \layerName{f.28} &  & $(512, 512)$ & & & $(14, 14)$ & & \\\hline

		\multirow{4}{*}{\mobilenet} & \multirow{4}{*}{\codeFont{mobilenet\_v2}}  & \layerName{f.8.c.2} & \multirow{4}{*}{$(1, 1)$} & $(384, 64)$ & \multirow{4}{*}{$(0, 0)$} & \multirow{4}{*}{$(1, 1)$} & $(14, 14)$ & \multirow{4}{*}{$0.71$} & \multirow{4}{*}{$0.70$} \\
		&& \layerName{f.11.c.2} &  & $(384, 96)$ & & & $(14, 14)$ & & \\
		&& \layerName{f.14.c.2} &  & $(576, 160)$ & & & $(7, 7)$ & & \\
		&& \layerName{f.17.c.2} &  & $(960, 320)$ & & & $(7, 7)$ & & \\\hline

		\multirow{4}{*}{\resnet} & \multirow{4}{*}{\codeFont{resnet50}}  & \layerName{l1.2.c3} & \multirow{4}{*}{$(1, 1)$} & $(64, 256)$ & \multirow{4}{*}{$(0, 0)$}  & \multirow{4}{*}{$(1, 1)$} & $(56, 56)$ & \multirow{4}{*}{$0.84$} & \multirow{4}{*}{$0.79$} \\
		&& \layerName{l2.3.c.3} &  & $(128, 512)$ & & & $(28, 28)$ & & \\
		&& \layerName{l3.5.c.3} &  & $(256, 1024)$ & & & $(14, 14)$ & & \\
		&& \layerName{l4.2.c.3} &  & $(512, 2048)$ & & & $(7, 7)$ & & \\\hline

		\multirow{3}{*}{\convnext} & \multirow{3}{*}{\codeFont{convnext\_small}}  & \layerName{f.2.1} & \multirow{3}{*}{$(2, 2)$} & $(96, 192)$ & \multirow{3}{*}{$(0, 0)$}  & \multirow{3}{*}{$(2, 2)$} & $(28, 28)$ & \multirow{3}{*}{$0.88$} & \multirow{3}{*}{$0.82$} \\
		&& \layerName{f.4.1} &  & $(192, 384)$ & & &  $(14, 14)$ & & \\
		&& \layerName{f.6.1} &  & $(384, 716)$ & & &  $(7, 7)$ & & \\\bottomrule

	\end{tabulary}
\end{table*}

%% file: tables/coreVecSizeRange.tex
\begin{table*}
	\centering
	\caption{Sum of $\coreVecSize$ (\corevector size) across the selected layers for tuning configurations with the highest $\aucGeomAll$, for \cifar and \imagenet. \avgpDimRed is constant across datasets and \gls{dmd}/\gls{macs}, corresponding to the sum of $\channelsOut$ of the selected layers (Table \ref{tab: layer selection}).}
	\label{tab: corevec size sum}
	\begin{tabulary}{\linewidth}{l|C|C|C|C|C|C|C|C|C}
		\toprule
		\multirow{3}{*}{Model} & \multirow{3}{*}{\avgpDimRed} & \multicolumn{4}{c|}{\cifar} & \multicolumn{4}{c}{\imagenet} \\\cline{3-6}\cline{7-10}
		& & \multicolumn{2}{c|}{\gls{dmd}} & \multicolumn{2}{c|}{\gls{macs}} & \multicolumn{2}{c|}{\gls{dmd}} & \multicolumn{2}{c}{\gls{macs}} \\\cline{3-4}\cline{5-6}\cline{7-8}\cline{9-10}
		& & \toeplitzDimRed & \kernelDimRed & \toeplitzDimRed & \kernelDimRed & \toeplitzDimRed & \kernelDimRed & \toeplitzDimRed & \kernelDimRed \\\midrule
		\vgg & $1408$ & $2254$ & $982$ & $2146$ & $812$ & $2687$ & $504$ & $3228$ & $865$ \\\hline
		\mobilenet & $640$ & $3336$ & $344$ & $2471$ & $568$ & $2147$ & $416$ & $3011$ & $395$ \\\hline
		\resnet & $3840$ & $1714$ & $859$ & $1714$ & $528$ & $1498$ & $366$ & $3120$ & $824$ \\\hline
		\convnext & $1292$ & $1556$ & $1052$ & $474$ & $908$ & $2421$ & $626$ & $2530$ & $689$ \\\bottomrule
	\end{tabulary}
\end{table*}

%% file: tables/footprint.tex
\begin{table*}[h!]
	\centering
	\caption{
		\gls{svd} memory usage and average $\coreVec$ computation time per $250$-samples batches when considering the maximum $\coreVecSize$ (memory decreases proportionally to $\coreVecSize$ for smaller $\coreVecSize$s), and average computation times and standard deviation for $250$-samples batches for \gls{dmd} and \gls{macs}. Averages correspond to $20$ batches for a total of $5000$ samples. The hyper-parameters used correspond to the best configurations with highest $\aucGeomAA$. Sizes are in Bytes and times are in $ms$.
	}
	\label{tab: mem and comp time}
	\begin{tabulary}{\linewidth}{L|L|C|C|C|C|C|C|C|C}
		\toprule
		\multicolumn{1}{c}{} & & \multirow{2}{*}{\gls{svd}\ Size} & \multirow{2}{*}{$\coreVec$ Time} & \multicolumn{3}{c|}{\cifar} & \multicolumn{3}{c}{\imagenet} \\\cline{5-7}\cline{8-10}
		\multicolumn{1}{c}{} & & & & $\coreVec$ Size & \gls{dmd}\ Time & \gls{macs}\ Time & $\coreVec$ Size & \gls{dmd}\ Time & \gls{macs}\ Time \\\midrule

		\multirow{3}{*}{\vgg} & \avgpDimRed & $4.8$G & $0.3\pm0.2$ & $666$M & $3.62\pm0.02$ & $0.43\pm0.02$ & $773$M & $6.89\pm0.02$ & $\mathbf{0.50\pm0.03}$ \\\cline{2-10}
		& \toeplitzDimRed & $-$ & $0.4\pm0.1$ & $1.9$G & $3.82\pm0.02$ & $0.45\pm0.04$ & $2.2$G & $7.10\pm0.02$ & $0.51\pm0.11$ \\\cline{2-10}
		& \kernelDimRed & $4.5$M & $0.3\pm0.1$ & $666$M & $\mathbf{3.53\pm0.01}$ & $\mathbf{0.39\pm0.03}$ & $773$M & $\mathbf{6.15\pm0.03}$ & $0.51\pm0.03$ \\\midrule

		\multirow{3}{*}{\mobilenet} & \avgpDimRed & $1.1$G & $0.3\pm0.3$ & $303$M & $\mathbf{1.27\pm0.01}$ & $0.44\pm0.03$ & $352$M & $\mathbf{3.27\pm0.01}$ & $0.67\pm0.11$ \\\cline{2-10}
		& \toeplitzDimRed & $-$ & $0.6\pm0.2$ & $1.9$G & $1.33\pm0.02$ & $0.49\pm0.03$ & $2.2$G & $3.35\pm0.05$ & $0.80\pm0.26$ \\\cline{2-10}
		& \kernelDimRed & $2.4$M & $0.3\pm0.2$ & $303$M & $1.30\pm0.03$ & $\mathbf{0.43\pm0.07}$ & $353$M & $3.37\pm0.04$ & $\mathbf{0.54\pm0.09}$ \\\midrule

		\multirow{3}{*}{\resnet} & \avgpDimRed & $7.2$G & $0.4\pm0.2$ & $1.8$G & $2.33\pm0.01$ & $0.58\pm0.05$ & $2.1$G & $6.96\pm0.01$ & $1.33\pm0.08$ \\\cline{2-10}
		& \toeplitzDimRed & $-$ & $0.8\pm0.3$ & $1.9$G & $2.20\pm0.01$ & $0.47\pm0.11$ & $2.2$G & $4.83\pm0.02$ & $0.63\pm0.20$ \\\cline{2-10}
		& \kernelDimRed & $6.7$M & $0.3\pm0.1$ & $454$M & $\mathbf{2.07\pm0.01}$ & $\mathbf{0.42\pm0.03}$ & $528$M & $\mathbf{4.63\pm0.03}$ & $\mathbf{0.59\pm0.09}$ \\\midrule

		\multirow{3}{*}{\convnext} & \avgpDimRed & $23$G & $0.3\pm0.2$ & $638$M & $7.92\pm0.03$ & $0.50\pm0.04$ & $738$M & $13.52\pm0.04$ & $0.69\pm0.07$ \\\cline{2-10}
		& \toeplitzDimRed & $-$ & $1.1\pm0.4$ & $1.5$G & $7.98\pm0.03$ & $0.49\pm0.03$ & $1.7$G & $13.60\pm0.05$ & $0.72\pm0.27$ \\\cline{2-10}
		& \kernelDimRed & $24$M & $0.2\pm0.1$ & $638$M & $\mathbf{7.91\pm0.02}$ & $\mathbf{0.47\pm0.05}$ & $738$M & $\mathbf{9.93\pm0.03}$ & $\mathbf{0.60\pm0.07}$ \\\bottomrule
	\end{tabulary}
\end{table*}

%% file: sections/conclusions.tex
\section{Conclusions}\label{sec: conclusions}

This paper presents a novel \gls{dr} method with a controllable high-compression level for \glspl{cl}' activations to be used with \gls{ood} and \gls{aa} detection methods.
The method builds upon \gls{dmd}'s and \gls{macs}'s \gls{dr} methods;
from the former, we leverage their intuition that \glspl{llf} are present across the layer's output channels, and that simple average pooling is an effective way to achieve high-compression levels;
from the latter, we leverage the trimmed \gls{svd}, which enables control over the compression level.
To connect both \glspl{dr}, and hence leverage their benefits, we propose using a \gls{cl} isomorphism studied in \cite{praggastis2022svd}, which leads to relatively small dense matrices.

We extend \gls{dmd}'s and \gls{macs}'s implementations to enable them to accept generic \glspl{dr}, and compare their performance with the different \gls{dr} methods on \gls{ood} and \gls{aa} detection.
Results over standard \gls{ood} datasets and \glspl{aa} show that the proposed method leads to detection \glspl{auc} comparable to or superior to their original ones, indicating that it captures the information contained within the layer's activations, despite its inherently higher compression level.
Furthermore, results show that the proposed method effectively reduces the memory and computation footprints of \gls{ood}/\gls{aa} detection methods, contributing to the their broader applicability.